\documentclass[10pt,twocolumn,letterpaper]{article}

\usepackage[cvpr]{cvpr}      % To produce the REVIEW version
\usepackage{graphicx}
\usepackage{amsmath}
\usepackage{multirow}
\usepackage{tabularx}
\usepackage{amssymb}
\usepackage{booktabs}
\usepackage{cuted}
\usepackage{xcolor}
\usepackage{tikz}
\usepackage{pifont}
\usepackage{algorithm}
\usepackage{algpseudocode}
\usepackage{xcolor}

\usepackage[pagebackref,breaklinks,colorlinks]{hyperref}

\usepackage[capitalize]{cleveref}
\crefname{section}{Sec.}{Secs.}
\Crefname{section}{Section}{Sections}
\Crefname{table}{Table}{Tables}
\crefname{table}{Tab.}{Tabs.}

\def\cvprPaperID{12559} % *** Enter the CVPR Paper ID here
\def\confName{CVPR}
\def\confYear{2025}

\begin{document}


\title{Inpainting is All You Need: A Diffusion-based Augmentation Method for Semi-supervised Medical Image Segmentation}

\author{Xinrong Hu\\
University of Notre Dame\\
{\tt\small xhu7@nd.edu}
\and
Yiyu Shi\\
University of Notre Dame\\
{\tt\small yshi4@nd.edu}
}

\maketitle

%%%%%%%%% ABSTRACT
\begin{abstract}
Collecting pixel-level labels for medical datasets can be a laborious and expensive process, and enhancing segmentation performance with a scarcity of labeled data is a crucial challenge. 
This work introduces AugPaint, a data augmentation framework that  utilizes inpainting to generate image-label pairs from limited labeled data.
AugPaint leverages latent diffusion models, known for their ability to generate high-quality in-domain images with low overhead, and adapts the sampling process for the inpainting task without need for retraining.
Specifically, given a pair of image and label mask, we crop the area labeled with the foreground and condition on it during reversed denoising process for every noise level.
Masked background area would gradually be filled in, and all generated images are paired with the label mask.
This approach ensures the accuracy of match between synthetic images and label masks, setting it apart from existing dataset generation methods.
The generated images serve as valuable supervision for training downstream segmentation models, effectively addressing the challenge of limited annotations. 
We conducted extensive evaluations of our data augmentation method on four public medical image segmentation datasets, including CT, MRI, and skin imaging.
Results across all datasets demonstrate that AugPaint outperforms state-of-the-art label-efficient methodologies, significantly improving segmentation performance.
% Observing that output of Transformer is a sequence of patches, each of which is a projection of a input local area into latent space, we propose to maximize agreenment between patch embeddings speparately instead of getting the global features that is average of all patches.

% Also, experiments on different model architectures and with different self-supervised learning frameworks further prove our method has good generalization and can easily plug and play. 

\end{abstract}

\begin{figure*}
    \includegraphics[width=\linewidth]{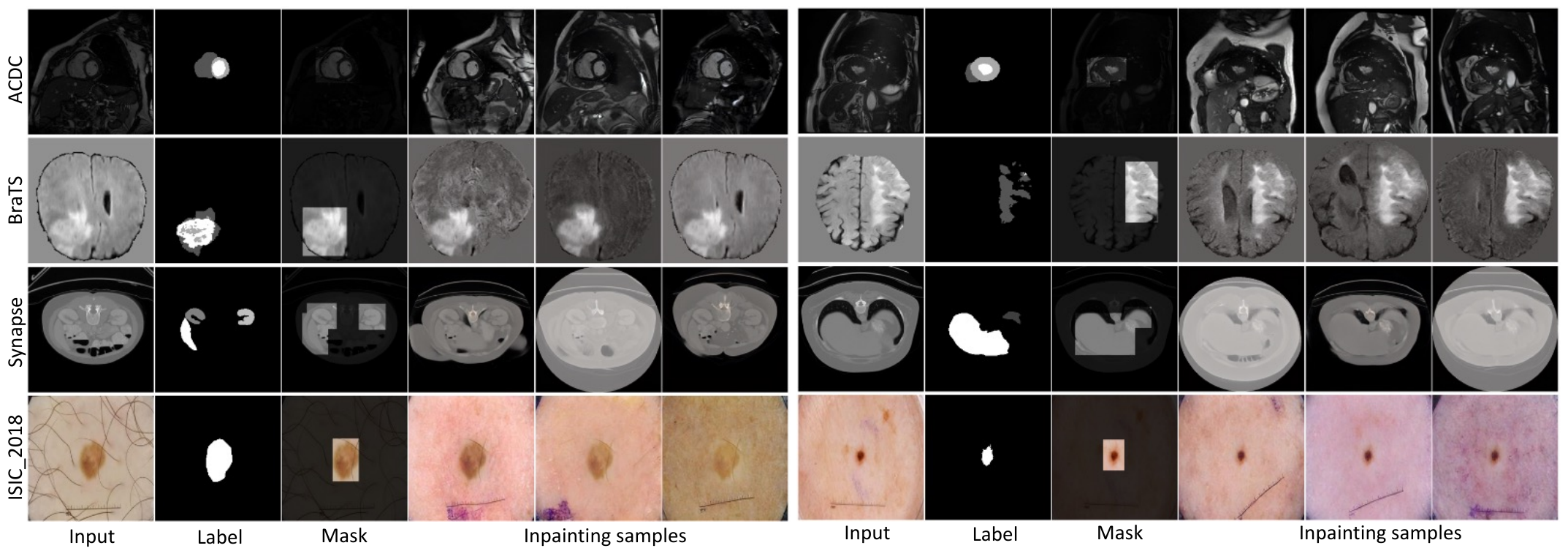}
    \caption{Samples generated with our AugPaint on four medical image segmentation datasets. Based on labeled data, we only keep boxes of labeled area and deploy latent diffusion model to inpaint masked background area. The generated images are paired with the provided label mask, and the dataset can be expanded to the desired extent. The generated backgrounds are well aligned with the input distribution and the contents exhibit semantically meaning.  }
    \label{fig:samples}
\end{figure*}

%%%%%%%%% BODY TEXT
\section{Introduction}

A substantial amount of labeled data is essential for training a segmentation model for medical images.
However, obtaining pixel-wise annotations is always time-consuming especially for high-resolution CT or MRI scans, and medical image labeling requires domain-specific professionals, which further leads to additional cost.
Moreover, the privacy issue of patients' data further limit the scale of public medical image dataset.
To address this problem, both self-supervised learning\cite{chen2020simple, chaitanya2020contrastive, hu2021semi} and semi-supervised learning\cite{sohn2020fixmatch, yang2023revisiting, yu2019uncertainty} methods dig into information undermined in unlabeled data, reducing the need of annotations.
%All these obstacles makes it compelling to augment dataset with limited number of labeled data.
On the other hand, generative methods introduce Generative Adversarial Network (GAN) to generate image-label pairs subject to the distribution of labeled data. 
However, GAN-based method either requires numerous image-label pairs during training\cite{neff2017generative, shin2018medical}, or generates images with inaccurate annotations \cite{zhang2021datasetgan, li2021semantic},
which has negative impacts on the finetuning stage.

In this work, we introduce inpainting, the process of filling in missing parts of an image, as a novel data augmentation method for generating additional image-label pairs in medical image segmentation (MIS).
Specifically, based on labeled data, we crop areas labeled as non-background and apply an inpainting model to filling in the background area, which ensure the match of inpainted image and the label of anchor image.
The motivation of using inpainting augmentation stems from two distinctive characteristics observed in MIS datasets:
1) The label mask is relatively small compared with natural dataset, which leaves more space for diverse image generation; 2) Foreground and background areas exhibit relatively small variations across images within the same dataset. Organs with the same label share similar shape and texture, enabling generated images to cover the distribution of the whole dataset and further to reduce overfitting.
Also, background contents of abdomen or brain scans show consistency across patients, and
less diverse background simplifies the inpainting task.

Recently, Denoising Probabilistic Diffusion Models (DDPM)\cite{dhariwal2021diffusion, ho2020denoising, nichol2021improved} emerges as a powerful image generation method showing superior performance over GAN models, and DDPM has also been applied for medical image generation\cite{akrout2023diffusion, ozbey2023unsupervised, pinaya2022brain}.
As for inpainting with DDPM, there are two primary research methodologies. The first, Palette\cite{saharia2022palette}, involves training a conditional diffusion model and then uses the unmasked area as the condition to sample image recovering contents in masked area.
Conversely, Repaint\cite{lugmayr2022repaint} is based on unconditional diffusion model and needs no retraining of diffusion model.
During inference, Repaint combines given pixels and sampled pixels at different denoising stages.
Although both of them exhibit promising performance in inpainting tasks, their extended sampling time inherent to DDPM imposes computational constraint for being a data augmentation method,

In this paper, we propose AugPaint, a novel diffusion-based image inpainting framework specifically designed to augment dataset in MIS. 
Inspired by latent diffusion model (LDM)\cite{rombach2022high}, AugPaint accelerates the sampling process by operating both adding noising process and reverse denoising process in latent space.
We compare two different inpainting methodologies mentioned above, with conditional LDMs and unconditional LDMs. 
Through analysis and experiment results, we uncover that inpainting with conditional LDM would transit to reconstructing original input image, which limits the diversity of generated samples.
Therefore, we claim that unconditional version of AugPaint is a better inpainting approach for data augmentation.
To validate the effectiveness of AugPaint, extensive experiments are conducted on four MIS datasets with various modalities and contents, including cardiac MRI, brain MRI, abdominal CT, and skin lesion images.
Under settings with constrained labeled data, results on all four datasets show that using synthetic image-label pairs generated by AugPaint for supervision outperforms both state-of-the-art self-supervised and semi-supervised methods.
Also, our synthetic dataset can be easily combined with other methods and show further performance boost.
%ControlNet\cite{zhang2023adding} imposes control by introducing an additional model that copies parameters from encoder and is connected to diffusion decoder via zero-convolution.

%Copy-Paste augmentation is a classic data augmentation method for image segmentation \cite{ghiasi2021simple}, which simply combine objects of interest of different labeled imgs.
\label{sec:intro}

%-------------------------------------------------------------------------
\section{Related Works}
 
\begin{figure*}
    \centering
    \includegraphics[width=\linewidth]{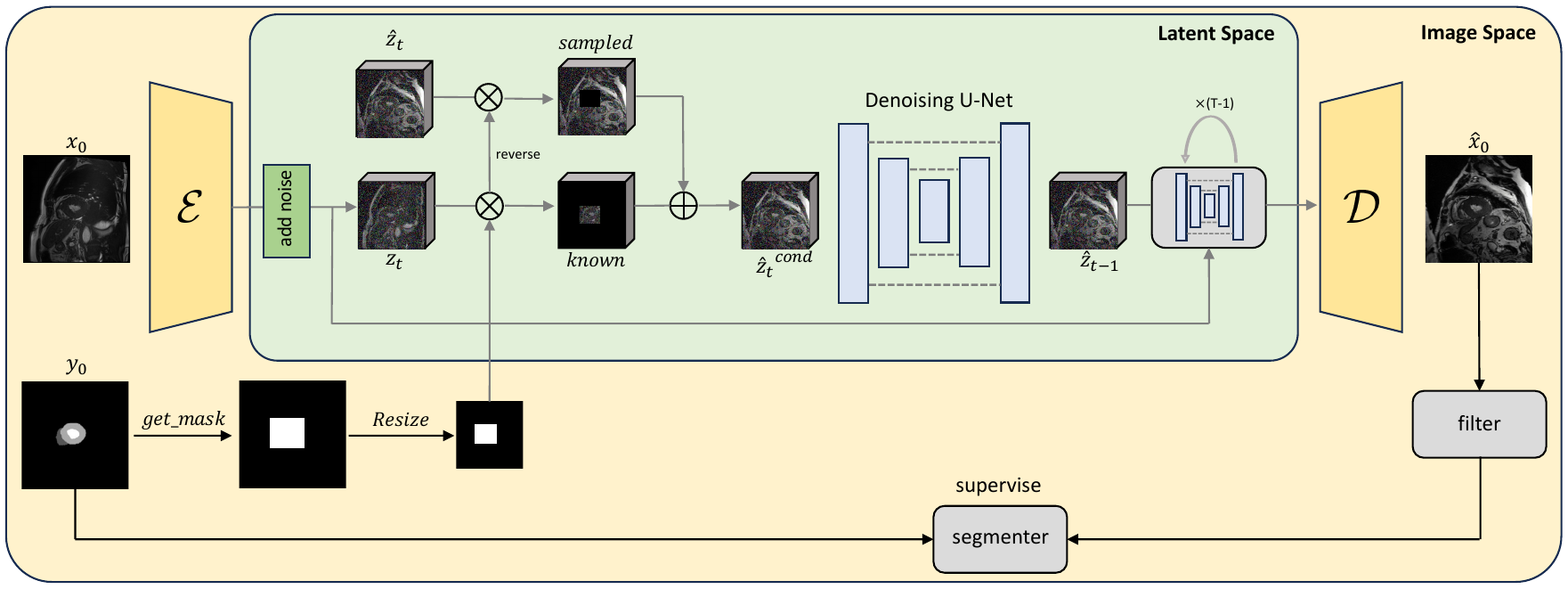}
    \caption{Illustration of sampling process via our AugPaint. Given a unconditional latent diffusion model trained on all data, we modify the denoising process to include the information in labeled area. The denoising process is operated in latent space for more efficient inference. }
    \label{fig:framework}
\end{figure*}

\textbf{Data augmentations for medical segmentation.} Besides transformations of original data\cite{nalepa2019data}, a range of deep learning models have been applied to synthesize training data to expand the training dataset\cite{kebaili2023deep}, including VAE (Variational Autoencoder)\cite{zhao2019data}, GAN(Generative Adversarial Network)\cite{neff2017generative, shin2018medical, mok2019learning, bailo2019red, shi2020novel}, and diffusion models\cite{khader2022medical, fernandez2022can, han2023medgen3d, du2023arsdm, yu2023diffusion}.
%GAN-based method requires pairs of image and label during training, and generated dataset fails to train a model with comparable segmentation accuracy trained on real dataset.
Alternatively, Zhao et al.\cite{zhao2019data} use learnable spatial and appearance transformations to generate more image-label pairs out of limited labeled data. 
Concurrently, Chaitanya et al\cite{chaitanya2021semi} deploy GAN as synthetic image generator to model shape and intensity variations.
However, the spatial transformation not only changes the shape of labeled area but also distorts background parts.
Hence, the synthetic images are out-of-distribution and adversely impact the following fine-tuning of segmentation model.

\textbf{Data augmentations with diffusion models.}
For natural images, several works exploit pretrained text-to-image diffusion models for classification task.
Trabucco et al.\cite{trabucco2023effective} edit image with an off-the-shelf diffusion model to increase semantic diversity while keeping the object corresponding to image-level annotation. 
He et al.\cite{he2022synthetic}explore if images generated by large text-to-image model can be used for improving classification and transfer learning.
To generate both images and pixel-level labels for segmentation task,
DiffuMask \cite{wu2023diffumask} utilizes Stable Diffusion (text-to-image generation diffusion model) and localizes class/word-specific regions with text-guided cross-attention information.
% maps between text and image pair to get accurate image-level annotations;
Dataset Diffusion \cite{nguyen2023dataset} shares the similar idea and further introduces class-prompt appending and self-attention exponentiation. 

In the field of MIS, the absence of large text-to-image diffusion models prompts different attempts. 
Khader et al.\cite{khader2022medical} generate high-quality 3D dataset with DDPM and acquires pseudo-label with segmentation model trained on another fully annotated dataset.
Fernandez et al.\cite{fernandez2022can} combine a diffusion-based label generator with a semantic image generator so that they can control the generation process with input segmentation label.
Another common strategy involves using label mask as condition for diffusion model to generate images with the identical label \cite{du2023arsdm, yu2023diffusion}.
However, with a limited number of image-label pairs, the synthetic image quality and the accuracy of matches between images and labels may be compromised.

\textbf{Label-efficient learning for MIS.} To address the challenge of annotation scarcity, self-supervised learning (SSL) has emerged as a promising approach. 
SSL formulates pretext tasks without the need of pixel-wise labels to learn general representation of datasets.
Contrastive learning, a key technique within SSL, has been adapted in various works to incorporate dataset-specific knowledge\cite{chaitanya2020contrastive, zeng2021positional, peng2021self, zhang2022unsupervised} or cater to task-specific objectives\cite{hu2021semi, you2022momentum, you2023bootstrapping}.
Another branch of works involves semi-supervised learning, which engages both labeled data and unlabeled data during training and usually follows a teacher-student framework. 
 Two typical techniques used to learn from unlabeled data include consistency regularization\cite{luo2022semi, bai2023bidirectional} and entropy minimization\cite{yu2019uncertainty, xia2020uncertainty, wu2022exploring}. 

%inpainting: \cite{saharia2022palette, rombach2022high} concatenate masked image $(img * (1-mask) + mask*noise\_like(img))$ with $x_{t}$ and only calculates the noise prediction loss in masked area.
%\cite{amit2021segdiff, baranchuk2021label} for natural image segmentation
%\cite{wolleb2022diffusion, rahman2023ambiguous} for medical images.
%\cite{wang2022semi}: Semi-Supervised Semantic Segmentation Using Unreliable Pseudo-Labels

\label{sec:related}

%-------------------------------------------------------------------------
\section{Method}

Let $x\in \mathbb{R}^{H\times W}$ be a 2D image or a slice of 3D volume and $y \in \mathbb{R}^{H\times W}$ be the corresponding pixel-wise label mask.
Within a given a medical image dataset $S$, we denote the subset with annotation as $S^l=\{(x^l, y^{l})\}$ and the remaining unlabeled data as $S^u=\{x^u\}$.
The objective of this work is to augment $S^l$ with additional synthetic pairs $(\hat{x}^{l}, \hat{y}^{l})$, then a larger labeled set can be used to train a segmentation model for better performance. 
To this end, we propose an inpainting framework AugPaint that relies on latent diffusion model to inpaint masked background of $x^{l}$ and keep unmasked area that contains all pixels labeled in $y^{l}$.
In the following sections, we firstly introduce implementing AugPaint with conditional LDM, and then detail the unconditional version of AugPaint.
Lastly, we discuss diversity issue caused by the conditional LDM and tackle the quality problem regarding the generated images with a filtering mechanism.

\subsection{Inpainting with Conditional LDM}
Inpainting process can be viewed as conditional generation.
To obtain a diffusion model specifically for inpainting task, we need to train a conditional LDM to learn the distribution $p(x|x^{mask})$ given the masked image $x^{mask}$.
% Latent diffusion firstly trains an autoencoder that serve as perceptual image compression.
For LDM, we firstly need to train an autoencoder that serves as a perceptual image compressor,   consisting of an encoder $\mathcal{E}$ and a decoder $\mathcal{D}$, 
The encoder $\mathcal{E}$ downsamples the input image $x_{0}$ by a factor of $f = 2^{m}$. 
We then have $z_{0} = \mathcal{E}(x_{0})$, transferring the pipeline from image space to latent space.
After randomly applying different noise level $t$, we can have a noisy image embedding $z_{t}$, where $t \in \{1, 2, ..., T\}$.
We use all datasets including $S^{l}$ and $S^{u}$ to train the conditional LDM.
As for the condition, since label information is inaccessible for all training data, we randomly generate a mask $x^{mask}$ and project it into latent space also with $E$, $z^{mask} = \mathcal{E}(x^{mask})$. 
To impose condition on $z_{t}$, we concatenate $z_{t}$ and $x_{m}$ as input of a noise predictor $\epsilon_{\theta}$ to predict the noise.
The objective loss can be simplified as 
\begin{equation}
    L_{c} := \mathbb{E}_{z_{t}, \epsilon \sim \mathcal{N}(0, I)}\left\| \epsilon - \epsilon_{\theta}(z_{t}, z^{mask}, t) \right\|
\end{equation}

During inference, we extract a binary mask from $y^{l}_{0}$, which contains bounding boxes of different segmentation classes, as shown in Fig \ref{fig:samples}. 
The intuition behind using bounding boxes instead of the direct label shape is twofold: Firstly, preserving the boundary of each class is crucial, because segmentation model learns by contrasting between the content within and outside the boundary; Secondly, using bounding boxes aids in maintaining the inherent shape of the target organ. Otherwise, during sampling, there is a potential risk of the model expanding the organ shape, which causes mismatch between synthetic samples and the input label. 
The condition mask $x^{mask}$ is the multiplication of the binary mask and input image $x_{0}$.
Starting from a Gaussian random noise $z_{T} \in \mathcal{N}(0, I)$, we can predict $\hat{z}_{t-1}| z^{mask}$ at every step $1 <= t <= T$ until we get $\hat{z}_{0}$. 
Then the decoder $\mathcal{D}$ upsamples $\hat{z}_{0}$ to full resolution $\hat{x}_{0}$.

\begin{figure}
    \centering
    \includegraphics[width=\linewidth]{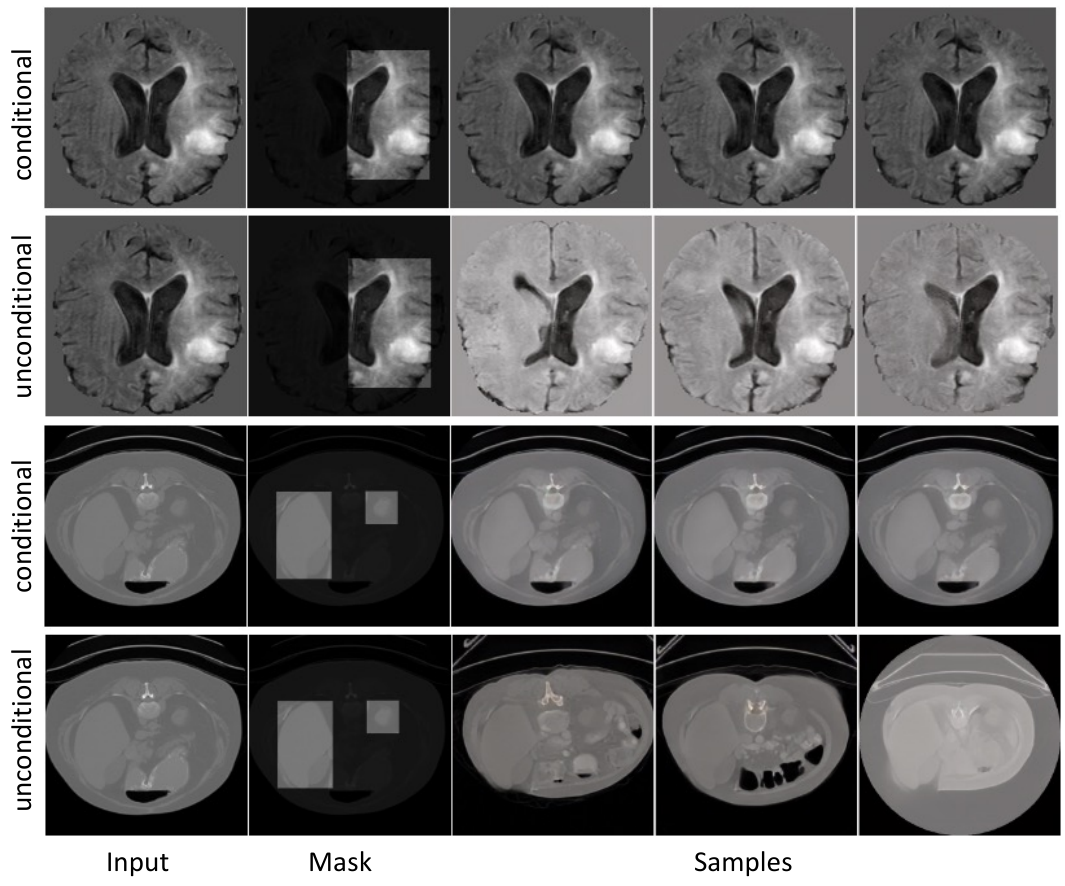}
    \caption{Comparisons of inpainting samples between with conditional LDM and unconditional LDM on BraTS and Synapse dataset. When the condition mask is large, conditional LDM tends to directly reconstruct the input image. In contrast, unconditional LDM can generate samples with better diversity regardless of condition mask size.  }
    \label{fig:diversity}
\end{figure}

\subsection{Inpainting with Unconditional LDM}
\label{sec:mask_type}
Inspired by Repaint\cite{lugmayr2022repaint}, we propose an approach leveraging an unconditional LDM for the task of inpainting, denoted as AugPaint.
This technique offers a significant advantage: it allows for the direct application of off-the-shelf pretrained LDMs, obviating the need for task-specific retraining, which is often resource-intensive.
As illustrated in Fig \ref{fig:framework}, the process begins with the projection of an input image $x_{0}$ into latent space.
The next step is the application of $q$ sample onto the downsampled image, resulting in $z_{t}$.
Concurrently, we extract a binary $mask$ from $y_{0}$ consisting of bounding boxes related to each segmentation class and subsequently resize this $mask$ to have the same dimensions of $z_{t}$.
At each iteration of denoising, we multiply $mask$ with $z_{t}$ targeting the known area of the image, and multiply the reversed $mask$ with the sampled $\hat{z}_{t}$ addressing the sampled area in latent space. 
Although the encoder $\mathcal{E}$ does not simply resize the image manifold, our empirical results demonstrate that the resized $mask$ aligns effectively with corresponding areas in $z_{t}$, which ensures that labeled areas from the original input $x_{0}$ are accurately preserved in the latent representation.
Combining the known area and the sampled area yields $\hat{z}^{cond}_{t}$, and subsequently the denoising U-Net $\epsilon_{\theta}$ predicts $z_{t-1}$ out of $\hat{z}^{cond}_{t}$.
For different $t$, we can always enhance the fidelity of the unmasked area with the ground truth in $x_{0}$.
The labeled area actually still serves as a condition but in a less apparent way during inpainting process. 
We further accelerate the sampling by applying DDIM\cite{song2020denoising}, and the sampling steps $T$ can be significantly reduced.
The detailed workflow of our AugPaint can be referred to Algorithm \ref{alg:inpaint}.
Thus the synthetic pairs $(\hat{x}_{0}, y_{0})$ generated by AugPaint serve as valuable supervisions for the segmentation model training.

\begin{algorithm}
\caption{Inpainting with our AugPaint based on unconditional LDM}\label{alg:inpaint}
\begin{algorithmic}
\State Input: input image $x_{0}$ with label $y_{0}$, sampling steps T, noise parameter $\{\alpha_{i}\}_{0:T} $
\State Output: inpainting sample $\hat{x}_{0}$
\State $z_{0} = \mathcal{E}(x_{0})$
\State $\hat{z}_{T} \sim \mathcal{N}(0, I)$
\For{\textbf{all} t from T to 1}
    \State $z_{t} = \sqrt{\alpha_{t}}z_{0} + (1 - \alpha_{t}) \epsilon$, with $\epsilon \sim  \mathcal{N}(0, I)$
    \State $mask = Resize(get\_mask(y_{0}))$
    \State $\hat{z}^{cond}_{t} = mask \odot z_{t} + (1 - mask) \odot \hat{z}_{t}$
    \State $\hat{\epsilon} = \epsilon_{\theta}(\hat{z}^{cond}_{t}, t)$
    \State $\tau_{t} = \sqrt{\frac{1-\alpha_{t-1}}{1-\alpha_{t}}}\sqrt{1- \frac{\alpha_{t}}{\alpha_{t-1}}}$
    \State $\bar{z}_{0} = \sqrt{\alpha_{t-1}}(\frac{x_{t} -\sqrt{1 - \alpha_{t}}\hat{\epsilon} }{\sqrt{\alpha_{t}}})$
    \State $\hat{z}_{t-1} = \bar{z}_{0} + \sqrt{1 - \alpha_{t-1} - \tau_{t}^{2}}\hat{\epsilon} + \tau_{t}\beta$, with $\beta \sim  \mathcal{N}(0, I)$
    \EndFor
\State $\hat{x}_{0} = \mathcal{D}(\hat{z}_{0})$
\State \textbf{return} $\hat{x}_{0}$
\end{algorithmic}
\end{algorithm}

\newcolumntype{Y}{>{\centering\arraybackslash}X}
\begin{table*}[t]
    \small
    %\footnotesize
    %\scriptsize
    \centering
    \caption{Comparisons between state-of-the-art methods and the proposed methods w.r.t. segmentation dice scores on ACDC dataset and BraTS dataset, using different number of labeled cases. All experiments are conducted on three independent splits. The highest dice score is highlighted, with underscoring indicating the best performance excluding AugPaint. A ``-" denotes dice score close to 0.}
    \label{table1}
    \begin{tabularx}{\linewidth}{c|YYY|YYY}
    \toprule
 \multicolumn{1}{c|}{\multirow{2}{*}{Methods}} & \multicolumn{3}{c|}{ACDC} & \multicolumn{3}{c}{BraTS} \\ \cline{2-7}
% & $0.05*|X_{tr}|$  & $0.1*|X_{tr}|$ & $0.2*|X_{tr}|$ & $0.1*|X_{tr}|$  & $0.2*|X_{tr}|$ & $0.4*|X_{tr}|$  \\
 &\multicolumn{1}{c}{1 case} & \multicolumn{1}{c}{2 cases} & \multicolumn{1}{c|}{5 cases}  & \multicolumn{1}{c}{1 case} & \multicolumn{1}{c}{2 cases} & \multicolumn{1}{c}{5 cases} \\
 \hline

U-Net &20.54$\pm$5.98 &31.33$\pm$12.35 &52.24$\pm$3.41 & 25.69$\pm$2.95 & 34.59$\pm$6.10 & 37.97$\pm$7.70 \\ 
Global\&Local \cite{chaitanya2020contrastive} &25.40$\pm$3.95 &35.41$\pm$2.20 &58.33$\pm$5.97 & 29.94$\pm$9.04 & 34.27$\pm$3.28 & 37.64$\pm$5.03 \\
PCL \cite{zeng2021positional} &27.64$\pm$3.37 &34.64$\pm$3.37 &59.52$\pm$1.83 & \underline{31.92$\pm$9.03} & 35.53$\pm$7.89 & 37.21$\pm$4.34 \\
\hline 
CNN\&Trans\cite{luo2022semi} & 34.91$\pm$5.32 &\underline{39.69$\pm$3.17} &60.23$\pm$7.62 &26.24$\pm$8.68  &33.97$\pm$8.14 & 38.05 $\pm$6.73 \\
SS-Net \cite{wu2022exploring} &19.28$\pm$8.62 &38.29$\pm$7.07 &63.70$\pm$9.46 &- &- &17.30$\pm$6.25  \\
UniMatch\cite{yang2023revisiting} &\underline{35.48$\pm$5.21} & 39.23$\pm$8.63 &\underline{64.56$\pm$6.87} &28.09$\pm$8.33 &\underline{35.87$\pm$7.50} &38.26$\pm$4.97 \\
\hline
Rand Aug &24.67$\pm$7.47  &31.85$\pm$10.2 &61.05$\pm$10.17 &30.10$\pm$9.75 &33.39$\pm$13.24 &\underline{38.45$\pm$2.85}  \\
ControlNet \cite{zhang2023adding} &32.64$\pm$8.62 &33.95$\pm$7.07 &50.64$\pm$9.46 &24.46$\pm$10.27  &31.58$\pm$11.44 &33.40$\pm$8.39 \\
cond AugPaint (\textbf{ours}) &38.67$\pm$10.93 &42.77$\pm$7.80 &74.05$\pm$6.62 & 32.53$\pm$ 9.46 & 38.83$\pm$4.08 &39.35$\pm$6.07
 \\
AugPaint(\textbf{ours}) &\textbf{42.42$\pm$9.60} &\textbf{46.15$\pm$6.98} &\textbf{77.73$\pm$4.21} & \textbf{35.19$\pm$7.96} &\textbf{39.60$\pm$5.40} & \textbf{41.85$\pm$5.53} \\ \hline
fully supervised & \multicolumn{3}{c|}{92.03$\pm$3.36 (w/ \textbf{70 cases}) } & \multicolumn{3}{c}{65.31$\pm$4.17 (w/ \textbf{414 cases})} \\
% \midrule[1pt]
% \multicolumn{1}{c}{benchmark} & \multicolumn{3}{c}{  0.878  (full $|X_{tr}|$)}                        & \multicolumn{3}{l}{}      \\
\bottomrule
\end{tabularx}
    
\end{table*}

\subsection{Sampling Diversity and Quality}
While both conditional and unconditional versions of AugPaint are capable of generating as many $\hat{x}_{0}$ as one would like from a single pair $(x_{0}, y_{0})$, the diversity of the resulting augmented dataset is crucial for the performance of segmentation models trained on it.
The training of LDM encompasses the entire dataset, including both labeled data and unlabeled data, aiming to synthesize images that represent the comprehensive distribution $p(x)$ of the whole dataset.
However, it is important to notice that the conditional LDM essentially learns $p(x|x^{mask})$, meaning the output is conditioned on the input mask $x^{mask}$.
In cases where the condition $x^{mask}$ is highly unique, the distribution $p(x|x^{mask})$ can directly equal $\{x_{0}\}$, transiting the inpainting process to a reconstruction task.
The resulting augmented dataset just contains replications of labeled data, which can cause overfitting during segmentation training phase, as shown in Fig \ref{fig:diversity}.
This usually happens when $x^{mask}$ is excessively large or the dataset exhibits significant variability across instances.
This issue is actually the same problem discussed in existing works \cite{carlini2023extracting, somepalli2023diffusion}, in which they extract training data from a pretrained diffusion model in test stage with a similar prompt used during training phase.
In contrast, the unconditional LDM predicts the unknown pixels in conjunction with known pixels and unknown pixels from previous iteration, rather than relying on an additional conditional input channel.
The condition is imposed in a more subtle and indirect manner, making it more resistant to this diversity issue.
In the experiment section, the results further prove the advantage of unconditional LDM quantitatively.

On the other hand, the quality of augmented dataset is equally pivotal in determining the success of training the segmentation model.
There are two potential problems involving quality assessment: deviation from original distribution $p(x)$ and mismatch between $\hat{x}_{0}$ and $y_{0}$. 
These quality issues usually stem from an overly small condition mask, which provides limited information for the inpainting process.
%thus making it more challenging to generate harmonic images.
On the other hand, small condition mask could lead to filling in organs of interests in the background area.
To address these issues, we adopt a two-step filtering strategy. 
We firstly employ the labeled data to train a segmentation model.
This model is then used to calculate the confidence score within the labeled areas on all generated images.
We sort all generated images by the confidence score and retain only the top half images.
% \begin{equation}

\begin{table*}[t]
    \small
    %\footnotesize
    %\scriptsize
    \centering
    \caption{Comparisons between state-of-the-art methods and the proposed methods w.r.t. segmentation dice scores on Synapse dataset and ISIC dataset, using different number of labeled cases. All experiments are conducted on three independent splits and we log the mean dice score and standard deviation. ``N/A" means PCL requiring 3D information is not applicable to skin lesion images.}
    \label{table2}
    \begin{tabularx}{\linewidth}{c|YYY|YYY}
    \toprule
 \multicolumn{1}{c|}{\multirow{2}{*}{Methods}} & \multicolumn{3}{c|}{Synapse}  & \multicolumn{3}{c}{ISIC\_2018} \\ \cline{2-7}
% & $0.05*|X_{tr}|$  & $0.1*|X_{tr}|$ & $0.2*|X_{tr}|$ & $0.1*|X_{tr}|$  & $0.2*|X_{tr}|$ & $0.4*|X_{tr}|$  \\
 &\multicolumn{1}{c}{1 case} & \multicolumn{1}{c}{2 cases} & \multicolumn{1}{c|}{5 cases}  & \multicolumn{1}{c}{10 cases} & \multicolumn{1}{c}{20 cases} & \multicolumn{1}{c}{50 cases} \\
 \hline

U-Net  & 21.07$\pm$7.38 & 24.79$\pm$9.61 & 55.51$\pm$3.89 &71.87$\pm$4.72 &75.56$\pm$1.76 &78.45$\pm$1.75 \\ 
Global\&Local \cite{chaitanya2020contrastive}&\underline{26.25$\pm$4.74} &32.31$\pm$7.62 & 56.72$\pm$5.23 &72.74$\pm$2.64 &76.32$\pm$5.39 &78.51$\pm$2.50 \\
PCL \cite{zeng2021positional} &24.82$\pm$3.98 &29.46$\pm$11.12 &57.14$\pm$3.48 & N/A & N/A & N/A \\ \hline
CNN\&Trans\cite{luo2022semi} &23.80$\pm$6.33 &25.18$\pm$7.91 &42.52$\pm$10.43  &70.84$\pm$4.13  &72.09$\pm$3.60 &76.04$\pm$2.42 \\
SS-Net \cite{wu2022exploring} &10.10$\pm$3.28  &27.41$\pm$8.15 &48.45$\pm$8.54 & 54.09$\pm$6.36 &59.35$\pm$6.19 &61.03$\pm$8.04 \\
UniMatch\cite{yang2023revisiting} & 27.72$\pm$7.17 & \underline{38.58$\pm$8.52} & 58.71$\pm$6.33 &\underline{73.18$\pm$4.39} &\underline{77.30$\pm$3.05} &\underline{79.87$\pm$2.41} \\
\hline
Rand Aug &21.81$\pm$7.97 &34.37$\pm$5.44 & \underline{60.28$\pm$2.38} &72.91$\pm$2.46 &76.68$\pm$0.64 &79.53$\pm$0.74 \\
ControlNet \cite{zhang2023adding} &16.81$\pm$5.99 &23.69$\pm$8.43 &41.08$\pm$7.54 & 71.84$\pm$5.60 &70.94$\pm$7.34 &71.99$\pm$5.51 \\
cond AugPaint (\textbf{ours}) & 34.53$\pm$ 9.46 & 49.03$\pm$9.70 & 62.37$\pm$4.42 &73.05$\pm$4.39 &79.57$\pm$1.16 &81.45$\pm$0.59 \\
AugPaint(\textbf{ours}) & \textbf{40.83$\pm$7.23} & \textbf{55.34$\pm$3.88} & \textbf{68.40$\pm$3.06} & \textbf{77.58$\pm$2.03} & \textbf{80.46$\pm$1.12} & \textbf{82.14$\pm$0.83} \\ \hline
fully supervised & \multicolumn{3}{c|}{73.99$\pm$2.14 (w/ \textbf{15 cases})} & \multicolumn{3}{c}{84.29$\pm$1.44 (w/ \textbf{2075 cases})} \\
% \midrule[1pt]
% \multicolumn{1}{c}{benchmark} & \multicolumn{3}{c}{  0.878  (full $|X_{tr}|$)}                        & \multicolumn{3}{l}{}      \\
\bottomrule
\end{tabularx}
    
\end{table*}

\label{sec:method}

%-------------------------------------------------------------------------
\section{Experiments}
\subsection{Dataset}
\textbf{ACDC.} The Automated Cardiac Diagnosis Challenge\cite{bernard2018deep} dataset is part of the MICCAI 2017 challenges, which contains MRI scans of cardiac structures from 100 patients, each with two 3D volumes. This dataset also provides expert segmentation masks of the left ventricle, right ventricle, and myocardium. 
We randomly split the MRI scans on patient basis into three parts, training set, validation set, and test set, with ratio being 70:15:15. 

\textbf{Brain Tumor Segmentation.}
This dataset is one of the segmentation tasks from Medical Segmentation Decathlon dataset\cite{antonelli2022medical}, and the task contains 484 MRI images with annotations of three types of tumor regions, which are namely the whole tumor (WT), enhancing tumor (ET), and tumor core (TC), respectively.
For every MRI, there are four channels representing different MR modalities, and we extract only the FLAIR channel.
We follow the data split in UNETR\cite{hatamizadeh2022unetr}, dividing the 484 images into three parts, train/validation/test, with ratio being 90:5:5.
% For evaluation, we report the dice sore and HD95 of three types of tumor, namely the whole tumor (WT), enhancing tumor (ET), and tumor core (TC).

\textbf{Synapse Multi-organ Segmentation.} 
% We evaluate the proposed UDA segmentation method on a abdominal multi-organ segmentation dataset with different modalities.
Synapse is from the \emph{Multi-Atlas Labeling Beyond the Cranial Vault Challenge}\cite{multi-atlas}, containing 30 volumes of CT scans. 
We focus on four organs with pixel-wise annotations including spleen, right kidney(RK), left kidney(LK), and liver.
The ratio of training set, validation set, and test set for each split is 2:1:1.

\textbf{Skin Lesion Segmentation.}
The 2018 \textbf{I}nternational \textbf{S}kin \textbf{I}maging \textbf{C}ollaboration (ISIC) Challenge \cite{codella2019skin} aims to perform semantic segmentation on colored dermoscopic images of skin lesions.
This dataset already provides a training set with 2594 labeled images and a validation set with 100 labeled images.
We split the provided training set into two parts 8:2 used for training and validation.
The test is operated on the provided validation set.

We convert all volumes in ACDC, BraTS, and Synapse to slices of 2D RGB images, thus all four datasets have the same input format.
The images are all reshaped to [256, 256, 3].
We independently generate three splits for each dataset, and repeat the experiments for three times.

\subsection{Implementation details}
The experiments are designed to tackle limited label problem in medical image segmentation. 
We randomly sample very few labeled data in the training set as $S^{l}$ and treat all other images as unlabeled set $S^{u}$.
We use both $S^{l}$ and $S^{u}$ for training LDM.
Since there are no pretrained LDM checkpoints online for the above four datasets, we need to train the LDM by ourselves.
We strictly follow the instructions in \cite{rombach2022high}, firstly training a VQ-regularized autoencoder with the compression rate $f=4$. 
Thus images in latent space have dimensions of 64x64.
Sequentially, we train a denoising U-Net with four resolution stages.
The batch sizes for both tasks are set as 4 and the training epochs are both 1000.
As for sampling, we adopt DDIM sampling with 50 steps to generate 5 inpainting samples for every pairs in $S^{l}$.
It takes average \textbf{5.42s} to generate 8 samples on a NVIDIA A10 gpu,  almost \textbf{20}$\times$ faster than RePaint\cite{lugmayr2022repaint} with 50 steps and 5 times resampling. 
The segmentation training only employs the augmented labeled dataset.
The default architecture of the segmentation model is also based on U-Net.
This U-Net has four resolution stages and the initial channel number is 64.
The batch size is set as 4 and the model is trained for 120 epochs.
The optimizer is Adam with learning rate being 0.0005. 

\subsection{Comparison with the State of the Arts}
\textbf{Baselines.}
We implement two self-supervised methods Global\&Local \cite{chaitanya2021semi} and PCL\cite{zeng2021positional}, as well as three semi-supervised methods CNN\&Trans\cite{luo2022semi}, SS-Net\cite{wu2022exploring}, and UniMatch\cite{yang2023revisiting}.
For self-supervised methods, we use both labeled and unlabeled data for pre-training and deploy finetuning only with labeled data.
For fair comparison, we change the segmentation model backbone in these methods to be the same as our default U-Net.
We also present the results of supervision with limited labeled data (denoted as ``U-Net'') and supervision with all training data.
On the other hand, we also include two data augmentation approaches, ``Rand Aug'' and ControlNet \cite{zhang2023adding}.
``Rand Aug'' means applying random transformations to augment the dataset by 5 times, and we also apply the same filtering approach to keep only 50\% data.
ControlNet\cite{zhang2023adding} is an emerging technique that imposes control by introducing an additional model that copies parameters from encoder in LDM and is connected to diffusion decoder via zero-convolution.
We use ControlNet to generate samples conditioned on the label masks. 

\begin{table}[t]
    \small
    %\footnotesize
    %\scriptsize
    \centering
    \caption{Results of combing our AugPaint with three label-efficient methodologies using different number of labeled cases.}
    \label{table3}
    \resizebox{\columnwidth}{!}{
    \begin{tabular}{cr|ccc}
    \toprule
\multirow{2}{*}{Dataset} &\multicolumn{1}{c|}{\multirow{2}{*}{Method}} & \multicolumn{3}{c}{Dice\%$\uparrow$} \\ \cline{3-5}
& & 1 case & 2 cases & 5 cases\\ \hline
\multirow{4}{*}{ACDC} &AugPaint &42.42 &44.15 &77.73 \\
&\textbf{+} Global\&Local\cite{chaitanya2020contrastive} &45.22 &48.26 &80.90 \\ 
&\textbf{+} PCL\cite{zeng2021positional} &45.79 &49.14 &81.66 \\
&\textbf{+} UniMatch\cite{yang2023revisiting} &46.22 &50.58 &80.49 \\ \hline

\bottomrule
\multirow{2}{*}{Dataset} &\multicolumn{1}{c|}{\multirow{2}{*}{Method}} & \multicolumn{3}{c}{Dice\%$\uparrow$} \\ \cline{3-5}
& & 10 cases & 20 cases & 50 cases\\ \hline
\multirow{4}{*}{ISIC\_2018} &AugPaint &77.58 &79.06 &82.14 \\
&\textbf{+} Global\&Local\cite{chaitanya2020contrastive} &78.97 &79.76 &86.62 \\ 
%&\textbf{+} PCL\cite{zeng2021positional} &N/A &N/A &N/A \\
&\textbf{+} UniMatch\cite{yang2023revisiting} &78.24 &80.17 &83.56\\
\bottomrule
\end{tabular}
}
\end{table}

Table \ref{table1} \& \ref{table2} present the comparisons between baselines and our methods across the four MIS datasets with different numbers of labeled cases. 
The average dice score improvements that AugPaint achieves over training from scratch are 20.40 for ACDC, 6.13 for BraTS, 21.07 for Synapse, and 4.77 for ISIC\_2018.
Notably, both versions of AugPaint consistently outperform other baselines, demonstrating the superior advantage of our method on addressing limited labeled data problems.
We also observe that semi-supervised learning methods work well on ACDC dataset but fail to produce equally promising performance on other datasets.
This difference can be due to that ACDC exhibits less variations across patient scans compared with other datasets, which reduces the gap between labeled data and unlabeled data.
% Actually, ACDC is the default dataset in most of semi-supervised learning works.
This pattern implies a potential limitation of semi-supervised methods in generalization, while our method  free of complex designs showcases a more robust performance on diverse datasets.
Additionally, it shows that with limited image-label pairs, ControlNet struggles to synthesize high-quality images given label masks that can be used to effectively augment dataset.

Another advantage of our method is that AugPaint can be easily combined with other label-efficient methods.
For self-supervised learning, we can use the augmented dataset for finetuning, and for semi-supervised learning, we can conduct training on both the augmented dataset and unlabeled dataset.
Table \ref{table3} shows that the performance of AugPaint can be further boosted when combined with other approaches.
%In all, our method gets the 

\begin{figure}
    \centering
    \includegraphics[width=\linewidth]{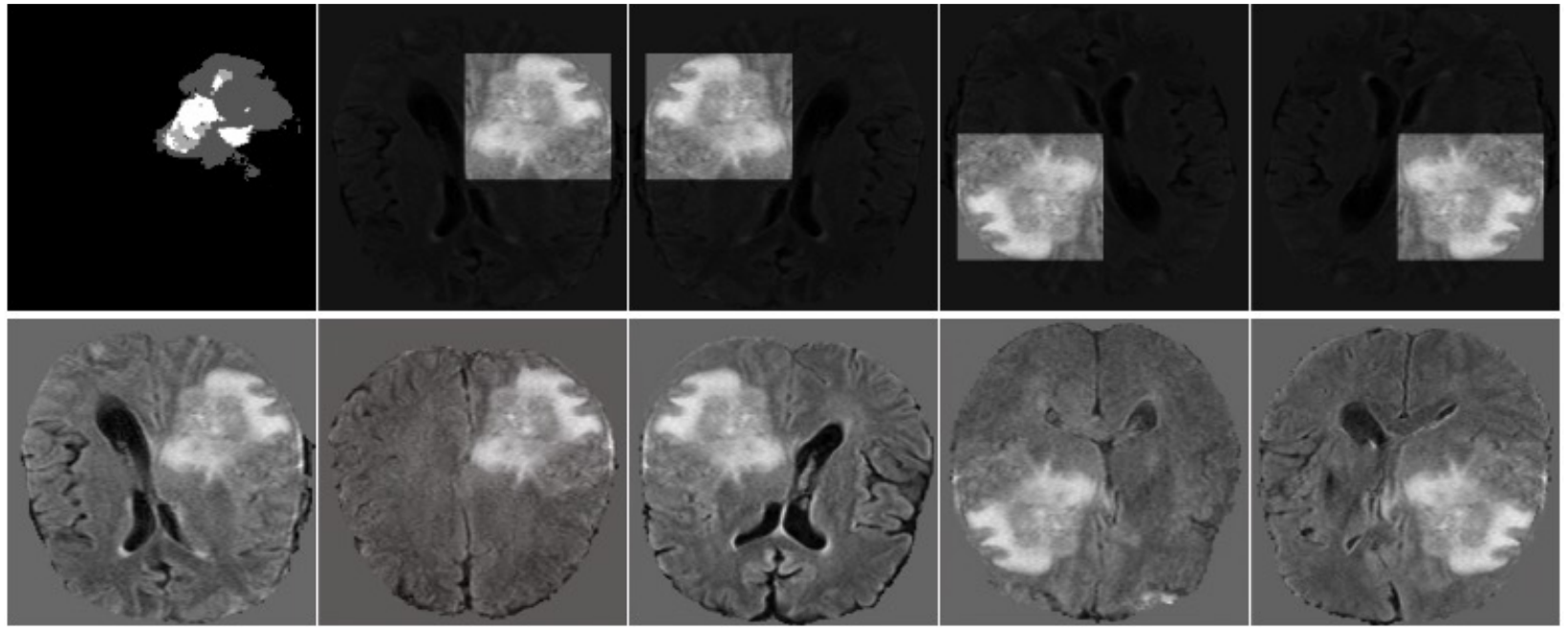}
    \caption{Inpainting with flipped masks on BraTS dataset. The first row is the label and flipped mask.
    The second row presents input image and generated inpainting samples corresponding to each mask}
    \label{fig:brats}
\end{figure}

\subsection{Mask Flipping for BraTS}
Unlike ACDC, Synapse, and ISIC\_2018 datasets, BraTS focuses on tumor segmentation with tumors potentially appearing anywhere within the brain scans.
In contrast, ventricles are supposed to be the center of cardiac scan in ACDC, and abdomen images have a fixed layout for different organs in Synapse.
To take advantage of this characteristic of BraTS dataset, we implement a specialized mask augmentation to further enhance the diversity of synthetic images.
In our approach, we extract the condition mask from input label and randomly flip the masked image in vertical and horizontal directions during the inpainting process, as illustrated in Fig \ref{fig:brats}.
It is important to notice that the sampled images are not merely direct flipped brain tumor scans.
This is because the LDM is trained with images without flipping or rotation.
This approach allows us to generate images with tumors located at various positions from a single patient.
The results in Table \ref{table4} quantitatively support the effectiveness of this  mask flipping method.
The improvement of segmentation accuracy is attributed to the enhanced diversity of the augmented dataset generate by AugPaint with mask flipping.

\begin{table}[t]
    \small
    %\footnotesize
    %\scriptsize
    \centering
    \caption{Analysis of using flipped mask on BraTS dataset using different number of labeled cases. We compare dice score and HD95 between AugPaint and AugPaint with flipped mask. }
    \label{table4}
    \resizebox{\columnwidth}{!}{
    \begin{tabular}{l|ll|ll}
    \toprule
\multirow{2}{*}{Method} &\multicolumn{2}{c|}{\multirow{1}{*}{1 case}} &\multicolumn{2}{c}{\multirow{1}{*}{5 cases}} \\ 
& Dice\%$\uparrow$ & HD95$\downarrow$ & Dice\%$\uparrow$ & HD95$\downarrow$ \\ \hline
AugPaint &35.19 &62.33 &41.85 &38.43 \\
+ flipping &36.27(\textbf{+1.08}) &48.38(\textbf{-13.95}) &43.11(\textbf{+1.26}) &35.22(\textbf{-3.21}) \\

\bottomrule
\end{tabular}
}
\end{table}

\subsection{Ablation Studies}
\label{sec:ablation}
\textbf{Model architecture}. Table \ref{table5} shows performance of our methods with different model architectures. 
Alongside the default U-Net, we incorporate two additional popular segmentation models into analysis, the DeepLabV3\cite{chen2017rethinking} and Transunet\cite{chen2021transunet}.
Both DeepLabV3 and Transunet are implemented with with ResNet50 backbones, adhering to their default hyperparameters.
The total number of parameters is 31M for U-Net, 39M for DeepLabV3, and 105M for Transunet.
The results demonstrate that training with dataset generated by AugPaint improve segmentation accuracy for all types of segmentation models.
This underscores the general applicability of our AugPaint under constrained labeled data.
While the performance gain observed with DeepLabV3 and Transunet is comparatively less pronounced than with U-Net, this is because of the already high baseline dice scores achieved by these models.
\begin{table}[t]
    %\small
    \footnotesize
    %\scriptsize
    \centering
    \caption{Evaluation of our methods on three different segmentation architectures. We use 5 labeled cases for ACDC and 50 labeled cases for ISIC\_2018.}
    \label{table5}
    \resizebox{\columnwidth}{!}{
    \begin{tabular}{c|cccc}
    \toprule
Dataset & Model &w/ AugPaint & Dice\% $\uparrow$ & HD95 $\downarrow$ \\
\hline
\multirow{6}{*}{ACDC} & U-Net  &\ding{55} &52.24 & 22.28\\
  &UNet &\ding{51} &77.73 &10.06 \\ 
 &DeepLabV3  &\ding{55} &81.24 &7.06 \\
 &DeepLabV3 &\ding{51} &86.63 &2.38 \\  
 &Transunet  &\ding{55} &72.01 &15.09 \\
 &Transunet&\ding{51} &80.98 &6.28 \\
\hline
\multirow{6}{*}{ISIC\_2018} & U-Net  &\ding{55} &78.45 & 40.41\\
  &UNet &\ding{51} &82.14 &27.06 \\ 
 &DeepLabV3  &\ding{55} &82.03 &27.39 \\
 &DeepLabV3 &\ding{51} &83.60 &25.00 \\  
 &Transunet  &\ding{55} &83.13 &26.56 \\
 &Transunet&\ding{51} &84.57 &23.77 \\
\bottomrule
\end{tabular}
}
\end{table}

\begin{figure}
    \centering
    \includegraphics[width=\linewidth]{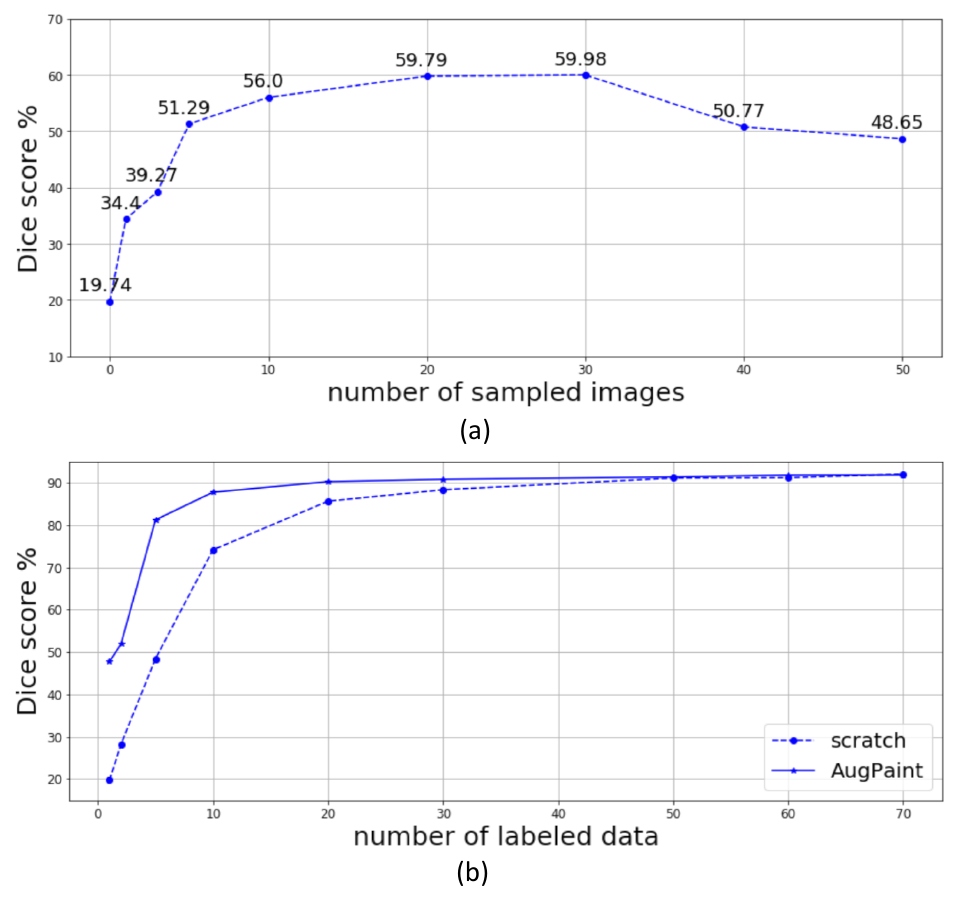}
    \caption{The influence of number of sampled images and number of labeled data on our AugPaint with ACDC dataset.  }
    \label{fig:num}
\end{figure}

\textbf{Number of samples}. In Fig \ref{fig:num} (a), we gradually increase the number of generated samples per labeled image and plot the segmentation metric of model trained with the augmented dataset.  
%Starting from the original labeled dataset without any augmentation, we augment the ACDC dataset with only 1 patient by up to $\times 50$ times.
As shown by the results, the performance is not consistently increasing with the addition of more sampled images.
The dice score peaks at around 30 sampled images per labeled image, and more synthetic data leads to performance decline beyond this point.
Our explanation for this observation is that when the number of sampled image becomes too large, our filtering process is less effective in removing all images of low-quality or with wrong label masks.
This leads to a contamination of the training dataset, adversely impacting model performance.
The decision of the number of samples per image should also take the sampling time into consideration, and generating 5 to 10 samples can be a good balance point between computational cost and segmentation accuracy. 
On the other hand, we explore the scenario of increasing the number of labeled data to assess if AugPaint still effectively improves segmentation performance when there are sufficient labeled data.
 Fig \ref{fig:num} (b) indicates that the advantage of using AugPaint is more significant with limited labeled data, and the dice score gap between our method and baseline gradually narrows with more labeled data.
 This is attributed to that we train the LDM solely with the available training set and without incorporating any additional data.
 Consequently, the augmented dataset generated by AugPaint cannot provide external information for training the segmentation model.
% We observe the same pattern on ISIC\_2018, more results can be seen in supplementary. 

\textbf{Masking type}. As discussed in sec \ref{sec:mask_type}, we prefer box-style masking over label-shape masking. 
We give two intuitions for this design. 
Table \ref{table6} further provides quantitatively support to the advantage of bounding box masking for inpainting.
More qualitative examples can be seen in supplementary. 
The results show that using label-shape mask on ISIC\_2018 is even worse than the baseline without any augmentations, which is caused by the mismatch between generated samples and labels.
In contrast, deploying bounding box masks is capable of reducing the chance of expanding organ shape outside labeled areas.

\begin{table}[t]
    %\small
    %\footnotesize
    \scriptsize
    \centering
    \caption{Ablation study of mask types used to obtain masked images. ``Label" means only keeping the labeled areas, while ``box" means extracting a bounding box surrounding the labeled areas.}
    \label{table6}
    \resizebox{\columnwidth}{!}{
    \begin{tabular}{cc|cc|cc}
    \toprule
\multirow{2}{*}{Dataset} &\multirow{2}{*}{Mask type} &\multicolumn{2}{c|}{\multirow{1}{*}{1(10) case}} &\multicolumn{2}{c}{\multirow{1}{*}{5(50) cases}} \\ 
& & Dice\%$\uparrow$ & HD95$\downarrow$ & Dice\%$\uparrow$ & HD95$\downarrow$ \\ \hline
\multirow{2}{*}{ACDC}&label &34.45 &43.87 &54.18 &20.63 \\
&box &42.42 &28.53 &77.73 &10.06 \\ \hline
\multirow{2}{*}{ISIC\_2018}&label &70.31 &63.24 &77.62 &42.41 \\
&box &77.58 &46.45 &82.14 &27.06 \\

\bottomrule
\end{tabular}
}
\end{table}

% \textbf{Filtering rate}

%n
\label{sec:exp}

%-------------------------------------------------------------------------

%source scans and target scans contain the same anatomical structures from the same view. The proposed method may not generalize to cases where 
%they are not. Potential mitigation include XXX. 
%Besides, even though the two datasets focus on the same objects, choosing slices in the 3D volumes from different anatomical planes would also cause performance degradation.
%Another limitation is that we train the segmentation model slice by slice, and the spatial information in the third dimension is discarded. 
%This is because there is no effective technique for volumes generation and style transfer between 3-D images currently.
%Moreover, our framework is in an end-to-end manner, thus we are forced to do 2D segmentation.
%Lastly, our 

\section{Conclusion}

In summary, this work develops AugPaint to address the critical challenge of improving segmentation performance in the medical image domain, particularly under the constraint with a scarcity of labeled data.
AugPaint stands out as an innovative data augmentation framework that employs latent diffusion model to inpaint unlabeled areas in latent space, which can efficiently generate synthetic images that are consistently aligned with input labels.
%We investigate two different implementations to condition the inpainting process, with the conditional LDM or with unconditional LDM.
We reveal that unconditional AugPaint implemented with unconditional LDM is capable of avoiding directly reconstructing the input images when the labeled area is excessively large, 
which ensures the diversity of the augmented dataset.
%This ensures the diversity of the augmented dataset and hence leads to better segmentation model training.
The efficacy of AugPaint is comprehensively validated on four medical image datasets, and our method significantly improves state-of-the-art self-supervised and semi-supervised approaches.
Moreover, the augmented dataset generated by AugPaint can be easily combined with other methods in a plug-and-play manner.

\textbf{Limitations} While AugPaint is a powerful tool for data augmentation in medical image segmentation, fundamental differences between medical and natural image datasets potentially hinder the broader application of AugPaint.
Firstly, a ``car'' can have different appearances within the same dataset, and inpainting cannot alter the shape and color of a ``car" in labeled data, potentially leading to overfitting.
Moreover, the diverse real-world scenes make generation of a harmonized image much more challenging.
Additionally, our filtering method is relatively rudimentary and simply relies on a model trained with few labeled data.
In the future, we will work on developing more sophisticated approaches for identifying our-of-distribution images and excluding images with organs of interests generated in the background.

%Without access to large dataset, pre-training with a pretext task is then the key to success of ViT on computer vision tasks.
%This work further proves this argument in the field of medical image analysis, and focuses on a more challenging task, volumetric semantic segmentation.
%Considering the limitation of biomedical data and characteristics of segmentation task, we propose to predict patch-wise representations from different augmented views of the same volume, rather than working on a single global representation for each augmentation.
%Moreover, we devise two strategies to address the issue of representation collapse: one is ``rotate-and-restore", and the other is modifying the denominator in contrastive loss.
%This simple yet effective method can be used with different Transformer models, either hierarchical or not, and with different self-supervised learning architectures. 
% nnFormer pre-trained with our SimPROT-W demonstrates the best performance on both multi-organ segmentation and brain tumor segmentation tasks, when compared with other pretext approaches.

%%%%%%%%% REFERENCES
{\small
\bibliographystyle{ieee_fullname}
\bibliography{egbib}
}

\end{document}